\def\year{2022}\relax
\documentclass[letterpaper]{article} 
\usepackage{aaai22}  
\usepackage{times}  
\usepackage{helvet}  
\usepackage{courier}  
\usepackage[hyphens]{url}  
\usepackage{graphicx} 
\usepackage{natbib}  
\usepackage{caption} 
\DeclareCaptionStyle{ruled}{labelfont=normalfont,labelsep=colon,strut=off} 
\usepackage{algorithm}
\usepackage{algorithmic}

\usepackage{newfloat}
\usepackage{listings}
\floatstyle{ruled}
\newfloat{listing}{tb}{lst}{}
\floatname{listing}{Listing}

\usepackage{amsmath}
\usepackage{amssymb}
\usepackage{xfrac}
\usepackage{multirow}
\usepackage{array}
\usepackage{booktabs}
\newcommand{\boldparagraph}[1]{\noindent{\bf #1} }
\DeclareMathOperator{\sigmoid}{sigmoid}

\usepackage[switch]{lineno}

\usepackage{color}

\title{Non-local Recurrent Regularization Networks for Multi-view Stereo}
\author{
    Qingshan Xu\textsuperscript{\rm 1}\thanks{Work done while Qingshan Xu was a visiting PhD student at ETH Zurich.}, Martin R. Oswald\textsuperscript{\rm 2}, 
    Wenbing Tao\textsuperscript{\rm 1}\thanks{Corresponding author.}, Marc Pollefeys\textsuperscript{\rm 2,4} and Zhaopeng Cui\textsuperscript{\rm 3}\\
}
\affiliations{
    \textsuperscript{\rm 1}National Key Laboratory of Science and Technology on Multispectral Information Processing,\\
    School of Artifical Intelligence and Automation, Huazhong University of Science and Technology, China\\

    \textsuperscript{\rm 2}Computer Vision and Geometry Group, ETH Zurich, Switzerland\\
    \textsuperscript{\rm 3}State Key Lab of CAD \& CG, Zhejiang University, China\\
    \textsuperscript{\rm 4}Microsoft Artificial Intelligence and Mixed Reality Lab, Zurich, Switzerland\\
}

\usepackage{bibentry}

\begin{document}
\maketitle

\begin{abstract}
In deep multi-view stereo networks, cost regularization is crucial to achieve accurate depth estimation. 
Since 3D cost volume filtering is usually memory-consuming, recurrent 2D cost map regularization has recently become popular and has shown great potential in reconstructing 3D models of different scales. 
However, existing recurrent methods only model the local dependencies in the depth domain, which greatly limits the capability of capturing the global scene context along the depth dimension. To tackle this limitation, we propose a novel non-local recurrent regularization network for multi-view stereo, named NR2-Net. Specifically, we design a depth attention module to capture non-local depth interactions within a sliding depth block. Then, the global scene context between different blocks is modeled in a gated recurrent manner. This way, the long-range dependencies along the depth dimension are captured to facilitate the cost regularization. Moreover, we design a dynamic depth map fusion strategy to improve the algorithm robustness. Our method achieves state-of-the-art reconstruction results on both DTU and Tanks and Temples datasets.
\end{abstract}

\section{Introduction}
%
Multi-view stereo (MVS) has been a hot research topic in computer vision for decades. It aims to recover the 3D geometry of a scene from a set of images with known camera parameters. Nowadays, multi-view stereo reconstruction is typically decomposed into two separate steps: depth map estimation and fusion~\cite{Galliani2015Massively,Schonberger2016Pixelwise,Xu2019Multi,Xu2020Learning,Xu2020Planar,Yao2018MVSNet,Gu2019Cas}. 
For these two steps, accurate depth map estimation is often challenging due to a variety of real-world problems, \emph{e.g.}, low-textured areas, thin structures, occlusions and reflective surfaces.

\begin{figure}[t]
	\centering
	\includegraphics[width=\columnwidth]{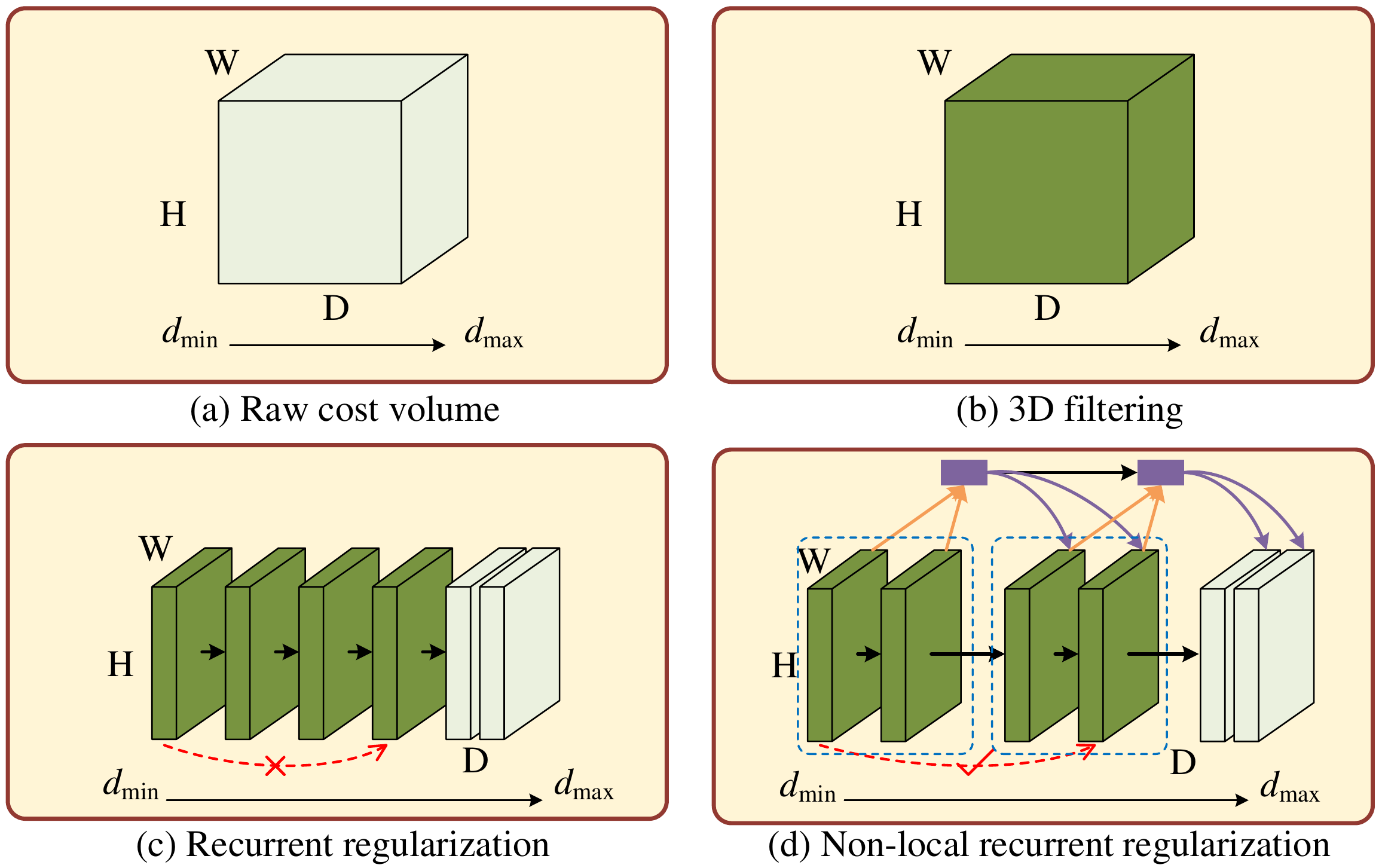}
	\caption{Illustration of different regularization methods. 3D filtering methods require cubic memory complexity while recurrent methods only have quadratic complexity.
	Our non-local recurrent regularization can explicitly establish interactions for non-adjacent depth values to capture more global context along the depth dimension which previous recurrent regularization methods cannot achieve.}
	\label{fig:teaser}
\end{figure}

Recently, many learning-based MVS methods~\cite{Yao2018MVSNet,Yao2019RMVSNet,Xu2020Learning,Gu2019Cas} have been presented to achieve competitive or better results than traditional methods that rely on optimization of photometric consistency~\cite{Furukawa2010Accurate,Schonberger2016Pixelwise}. 
Inspired by plane-sweeping stereo~\cite{Collins1996Space}, most learning-based methods first construct multi-channel cost volumes based on extracted deep image features, then perform deep cost volume processing to predict depth maps. 
The core of the successful learning-based MVS methods is the deep regularization of cost volumes to encode local and global information of a scene. 
The deep regularization converts the cost volume into a latent probability volume, directly influencing the final depth inference.

At present, the deep regularization of cost volumes can be categorized into two types: 3D cost volume filtering and recurrent 2D cost map regularization. The former one regularizes the entire 3D cost volume in one go, thus 3D convolutions can be seamlessly utilized to incorporate the context of a scene~\cite{Yao2018MVSNet,Xu2020Learning,Luo2019P}.
In general, this kind of method is memory-consuming since its memory requirement is cubic to the model resolution (cf. Fig~\ref{fig:teaser}(b)).
Although many follow-up works~\cite{Chen2019Point,Gu2019Cas,Cheng2020UCSNet,Yang2020CVPMVSNet} have been proposed to alleviate this problem, the scalability of 3D cost volume filtering is still limited when tackling high-resolution and large-scale 3D reconstruction.
As another line of research, the latter one~\cite{Yao2019RMVSNet,Yan2020Dense} sequentially regularizes 2D cost maps along the depth direction.
Therefore, this kind of method reduces the memory requirement from cubic to quadratic to the model resolution.
Moreover, such methods can adaptively sample sufficient depth hypotheses for scenes of different scales.
In order to aggregate spatial as well as temporal context information in the depth direction, recurrent networks, \emph{e.g.}, gated recurrent units (GRU) and long short-term memory (LSTM), are adopted in such methods.
However, these methods only explicitly consider the information interaction between adjacent depth values, hence the long-range dependencies along the depth dimension cannot be fully captured (cf. Fig~\ref{fig:teaser}(c)). 

To tackle the above limitation of recurrent 2D cost map regularization, we propose a novel non-local recurrent regularization network for multi-view stereo, namely NR2-Net.
Built upon the regular recurrent neural networks, we divide the depth sampling space into different blocks to model the non-local interactions for non-adjacent depth planes.
Within each block, we design a depth attention module to distill latent high-level features.
This module samples cost map features at every other depth planes to enable large receptive fields like dilated convolution~\cite{Chen2018DeepLab}.
Based on these cost map features, the latent high-level features are distilled via attention mechanism. 
Then the high-level features between blocks are further interacted in a gated recurrent manner to capture global scene context, which is used to regularize the bottom-level cost map features in the next block. 
In this way, the long-range dependencies along the depth dimension are modeled and the global scene context is perceived to help the long-range cost map regularization (cf. Fig~\ref{fig:teaser}(d)). 
At last, in order to fuse depth maps into point clouds, existing learning-based methods usually first predefine a constant depth probability threshold to filter out unreliable depth estimates. However, due to the uncertainty of depth prediction networks~\cite{Kendall2017Uncertainties}, this constant threshold will discard many credible depth values, resulting in point clouds of low completeness. 
Thus, we adaptively associate the depth probability threshold with depth consistency to generate more complete point clouds. 

Our contributions can be summarized as follows: 
\textbf{1)} We present a \textbf{novel non-local recurrent regularization} framework for multi-view stereo. This allows to perceive more global context information in the depth direction to assist cost volume regularization. 
\textbf{2)} We propose a \textbf{depth attention module} to distill latent high-level features along the depth dimension. This helps to model non-local interactions for non-adjacent depth hypotheses in the gated recurrent manner. 
\textbf{3)} We develop a \textbf{dynamic depth map fusion} strategy to reconstruct point clouds. This strategy jointly considers the depth probability and depth consistency in a dynamic way, which is robust for different scenes. Our method, NR2-Net, achieves state-of-the-art performance on both DTU and Tanks and Temples datasets.

\section{Related Work}
%
\boldparagraph{Traditional multi-view stereo.} In order to estimate depth maps for all input images, multi-view stereo needs to search correspondences across different images. Plane-sweeping methods~\cite{Collins1996Space,Gallup2007Real} sample depth plane hypotheses in the 3D scene. Then, they employ hand-crafted similarity metrics, \emph{e.g.}, sum of absolute differences (SAD) and normalized cross correlation (NCC), to construct cost volumes and extract the final depth maps via the winner-take-all strategy. Since these similarity metrics are ambiguous to some challenging areas, \emph{e.g.}, low-textured areas and reflective surfaces, some methods adopt engineered regularization technologies, such as graph-cuts~\cite{Kolmogorov2002Multi} and cost filtering~\cite{Hosni2013Fast}, to alleviate this problem. On the other hand, PatchMatch MVS methods~\cite{Zheng2014PatchMatch,Galliani2015Massively,Schonberger2016Pixelwise,Xu2019Multi,Xu2020Planar} adopt the sampling and propagation strategy~\cite{Barnes2009PatchMatch} to efficiently search continuous depth plane hypotheses from the whole depth interval. These methods impose implicit smoothness constraints based on the plane propagation in the 3D scene. Although they have greatly improved the performance of traditional methods, how to handle the ambiguity in the challenging areas is still an open problem. 

\begin{figure*}[t]
	\centering
	\includegraphics[width=0.95\linewidth]{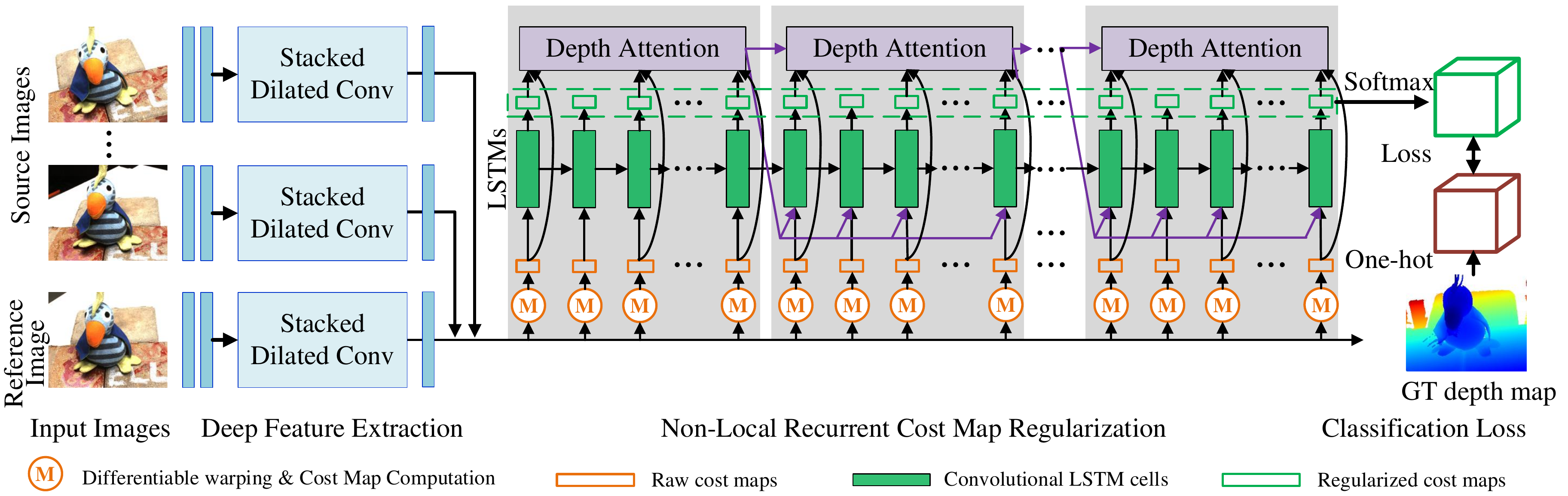}
	\caption{The NR2-Net architecture. Our network first extracts high-resolution deep features for all input images. Then cost maps at different depth planes are computed by differentiable homography warping and regularized by convolutional LSTM cells. To do non-local recurrent cost map regularization, the whole depth planes are divided into different blocks. The non-local interactions within each block are modeled by a depth attention module to distill latent high-level features. These high-level features are further interacted in a gated recurrent manner to capture global context information, which is used to help regularize the hidden states of LSTM cells in the next depth block.}
	\label{fig:NR2}
\end{figure*} 

\boldparagraph{Learning-based multi-view stereo.} 
Recently, some works have leveraged deep neural networks to learn deep similarity features and deep regularization for multi-view stereo~\cite{Hartmann2017Learned,Ji2017Surfacenet,Kar2017Learning}. 
MVSNet~\cite{Yao2018MVSNet} and DeepMVS~\cite{Huang2018DeepMVS} propose to construct cost volumes to learn depth maps for each input image.
This makes learning-based MVS methods more scalable and applicable to scene reconstruction. To infer depth maps, MVSNet and many following works~\cite{Xue2019MVSCRF,Luo2019P,Xu2020Learning} utilize multi-scale 3D convolutions to regularize cost volumes. However, due to the limitation of cost volume resolution, these methods still cannot tackle high-resolution images.
To predict high-resolution depth maps, cascade methods~\cite{Cheng2020UCSNet,Gu2019Cas,Yang2020CVPMVSNet} employ the coarse-to-fine framework to infer high-resolution estimation via thin cost volumes.
However, these cascade methods heavily rely on coarse-scale estimation which often lacks important detail information. 
Consequently, this may result in blurred effects in high-resolution estimation.
To achieve high-resolution prediction, R-MVSNet~\cite{Yao2019RMVSNet} recurrently regularizes cost maps through GRU cells.
This dramatically reduces memory consumption. 
However, the generic GRU cells do not allow the regularization to consider enough context information.
$\text{D}^2$HC-RMVSNet~\cite{Yan2020Dense} absorbs both the merits of LSTM and 2D U-Nets to present a 2D U-Net architecture with convolutional LSTM cells.
This architecture considers more context information in image domain and can directly operate on high-resolution image features.
In contrast to our work, existing recurrent methods only explicitly consider interactions between adjacent depth planes and thus lack global context along the depth dimension.

\section{Method}
%
The network architecture of our method, NR2-Net, is depicted in Fig.~\ref{fig:NR2}. 
The input to our network is a reference image $\boldsymbol{I}_0$ and $N-1$ source images $\{\boldsymbol{I}_i\}_{i=1}^{N-1}$, where $N$ is the total number of input images. 
Their camera intrinsic parameters $\{{\bf K}_i\}_{i=0}^{N-1}$ and relative extrinsic parameters $\{{\bf R}_i,{\bf t}_i\}_{i=0}^{N-1}$ are known. 
Our NR2-Net first extracts high-resolution deep features for all input images. 
The cost maps at different depth planes are computed by homography warping.  
Our non-local recurrent regularization framework is built upon convolutional LSTM cells to regularize 2D cost maps. 
To achieve non-local interactions for non-adjacent depth planes, we divide the whole depth interval into different blocks.
The different depth planes within each block are interacted by a depth attention module to distill latent high-level features. 
The global context information is captured by further interacting the high-level features in a gated recurrent manner. 
Then, the global context information is in turn used to regularize the hidden states of different LSTM cells in the next block. 
This helps each LSTM cell perceive more global context information in the depth direction. 
Finally, we convert these hidden states into regularized cost maps to predict the depth map of the reference image.

\subsection{Cost Map Construction}
%
Similar to previous works, we use the plane-sweeping algorithm~\cite{Collins1996Space} to construct cost maps at different depth planes.
Specifically, after extracting high-resolution deep image features $\{\boldsymbol{F}_{i}\}_{i=0}^{N-1}$ for all input images by stacked dilated convolutions~\cite{Yan2020Dense}, we compute the warped image features $\{\tilde{\boldsymbol F}_i(d)\}_{i=1}^{N-1}$ for $N-1$ source images at depth value $d$ by differentiable homography warping.
Then the 2D cost map at depth value $d$ is calculated by:
\begin{equation}
\boldsymbol{C}(d)=\frac{\sum_{i=1}^{N-1}\big(1+{\boldsymbol w}_i(d)\big)\odot\big(\tilde{\boldsymbol F}_i(d)-\boldsymbol{F}_0\big)^2}{N-1},
\end{equation} 
where ${\boldsymbol w}_i(d)$ is the view weight of the $i$-th source image and `$\odot$' represents element-wise multiplication.
Herein, ${\boldsymbol w}_i(d)$ is computed by applying one convolution layer followed by group normalization and rectified linear units, a residual block~\cite{He2016Deep} and one convolution layer followed by a sigmoid function on the feature difference, $\tilde{\boldsymbol F}_i(d)-\boldsymbol{F}_0$. 
In this way, a cost volume $\{\boldsymbol{C}(d)\}_{d=0}^{D-1}$ can be obtained by concatenating 2D cost maps of all depth planes in the depth direction, where $D$ is the total number of depth planes.

\subsection{Non-local Recurrent Regularization}
%
As the core of learning-based MVS methods, cost volume regularization is important to aggregate context information in both spatial and depth domain. 
Existing recurrent 2D cost map regularization methods cast the whole cost volume as a series of 2D cost maps in the depth direction and sequentially regularize 2D cost maps via stacked recurrent neural networks. 
Although such methods are memory-friendly, they only explicitly consider the local interactions between adjacent depth values.

\begin{figure}[t]
	\centering
	\includegraphics[width=\columnwidth]{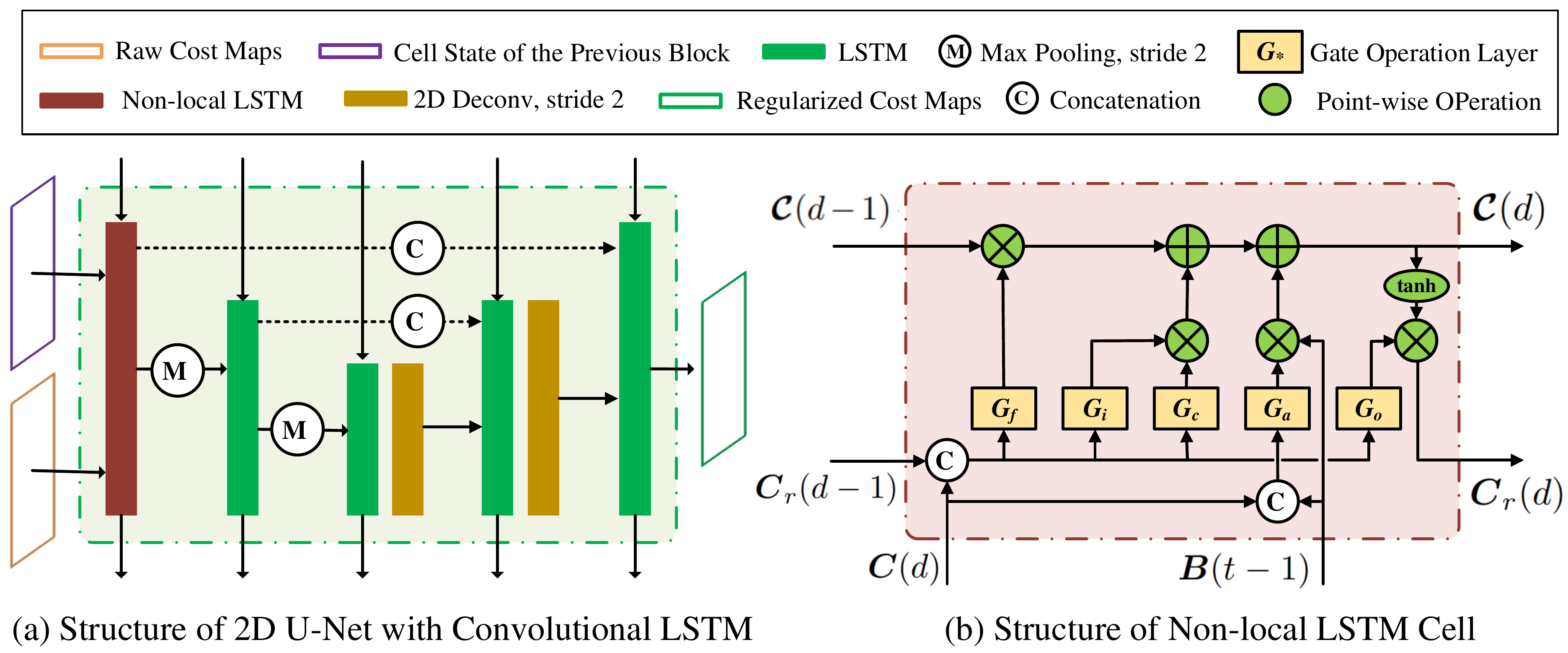}
	\caption{Illustrations of (a) structure of 2D U-Net with convolutional LSTM and (b) structure of non-local LSTM cell.}
	\label{fig:U-Net-LSTM}
\end{figure}

Inspired by non-local methods~\cite{Wang2018Non,Fu2019Non}, we design a non-local recurrent regularization network to model long-range dependencies along the depth dimension, which helps to regularize cost volumes by aggregating more context information. 
As illustrated in Fig.~\ref{fig:NR2}, we divide the whole cost volume into $T$ blocks, each block contains $s$ cost maps. 
Each cost map is regularized by stacked convolutional LSTM cells to incorporate context information in both the spatial and the depth domain.  
As depicted in Fig.~\ref{fig:U-Net-LSTM}(a), our stacked convolutional LSTM cells are built upon 2D U-Net architecture to aggregate multi-scale context information in the spatial domain. 
Moreover, these LSTM cells contain one non-local LSTM cell and four vanilla LSTM cells. 
This non-local LSTM cell is beneficial to incorporate more context information in the depth direction. 

Next we elaborate on the mechanism of our non-local LSTM cell. 
As illustrated in Fig.~\ref{fig:U-Net-LSTM}(b), besides using the current cost map $\boldsymbol{C}(d)$, the previous regularized cost map $\boldsymbol{C}_r(d-1)$ and the previous cell state map $\boldsymbol{\mathcal C}(d-1)$ as input like the vanilla LSTM cells, the non-local LSTM cell also incorporates the cell state map of the previous block, $\boldsymbol{B}(t-1)$, to consider long-range dependencies in the depth direction. 
How to obtain $\boldsymbol{B}(t-1)$ through the depth attention module will be detailed in the next section.
With these inputs, the forget gate map ${\bf F}(d)$, candidate state map $\hat{\boldsymbol{\mathcal C}}(d)$, input gate map ${\bf I}(d)$ and output gate map ${\bf O}(d)$ are respectively modeled as:
\begin{align}
{\bf F}(d)&=&\sigmoid&({\bf W}_{f}\ast[\boldsymbol{C}(d),\boldsymbol{C}_r(d-1)]+{\bf b}_f),\\
{\bf I}(d)&=&\sigmoid&({\bf W}_{i}\ast[\boldsymbol{C}(d),\boldsymbol{C}_r(d-1)]+{\bf b}_i),\\
\hat{\boldsymbol{\mathcal C}}(d)&=&\tanh&({\bf W}_{c}\ast[\boldsymbol{C}(d),\boldsymbol{C}_r(d-1)]+{\bf b}_c),\\
{\bf O}(d)&=&\sigmoid&({\bf W}_{o}\ast[\boldsymbol{C}(d),\boldsymbol{C}_r(d-1)]+{\bf b}_o),
\end{align}
where `$\ast$' means convolution, `$[]$' means concatenation, ${\bf W}$ is a transformation matrix and ${\bf b}$ is a bias term. Then, the current cell state map $\boldsymbol{\mathcal C}(d)$ and the current regularized cost map $\boldsymbol{C}_r(d)$ are computed as,
\begin{equation}\label{Eq:nl}
\boldsymbol{\mathcal C}(d)={\bf F}(d)\odot\boldsymbol{\mathcal C}(d-1)+{\bf I}(d)\odot\hat{\boldsymbol{\mathcal C}}(d)+{\bf A}(d)\odot\boldsymbol{B}(t-1),
\end{equation}
\begin{equation}
\boldsymbol{C}_r(d)={\bf O}(d)\odot\tanh(\boldsymbol{\mathcal C}(d)),
\end{equation}
where ${\bf A}(d)$ is the depth attention gate map given by
\begin{equation}
{\bf A}(d)=\sigmoid({\bf W}_{a}\ast[\boldsymbol{C}(d),\boldsymbol{B}(t-1)]+{\bf b}_a).
\end{equation}
Note that, the last term in the right side of Eq.~\eqref{Eq:nl} reflects the non-local interactions in the depth direction. This is the difference between our non-local LSTM cell and the vanilla LSTM cell.
Finally, the regularized cost maps $\{\boldsymbol{C}_r(d)\}_{d=0}^{D-1}$ pass a softmax layer to produce the probability volume $\boldsymbol{P}$. In this way, our regularization not only incorporates multi-scale context information in the spatial domain, but also aggregates long-range dependencies in the depth direction. 

\subsection{Depth Attention Module}
%
In the previous section, the depth attention module plays an important role in distilling latent high-level features within each block. 
It captures discriminative depth context information for each block. 
Then, the high-level features between blocks are interacted in a gated recurrent manner to capture global depth context information.
Finally, we can use this information to impose explicit long-range regularization constraints for non-adjacent depth values as described above.

\begin{figure}[t]
	\centering
	\includegraphics[width=0.95\linewidth]{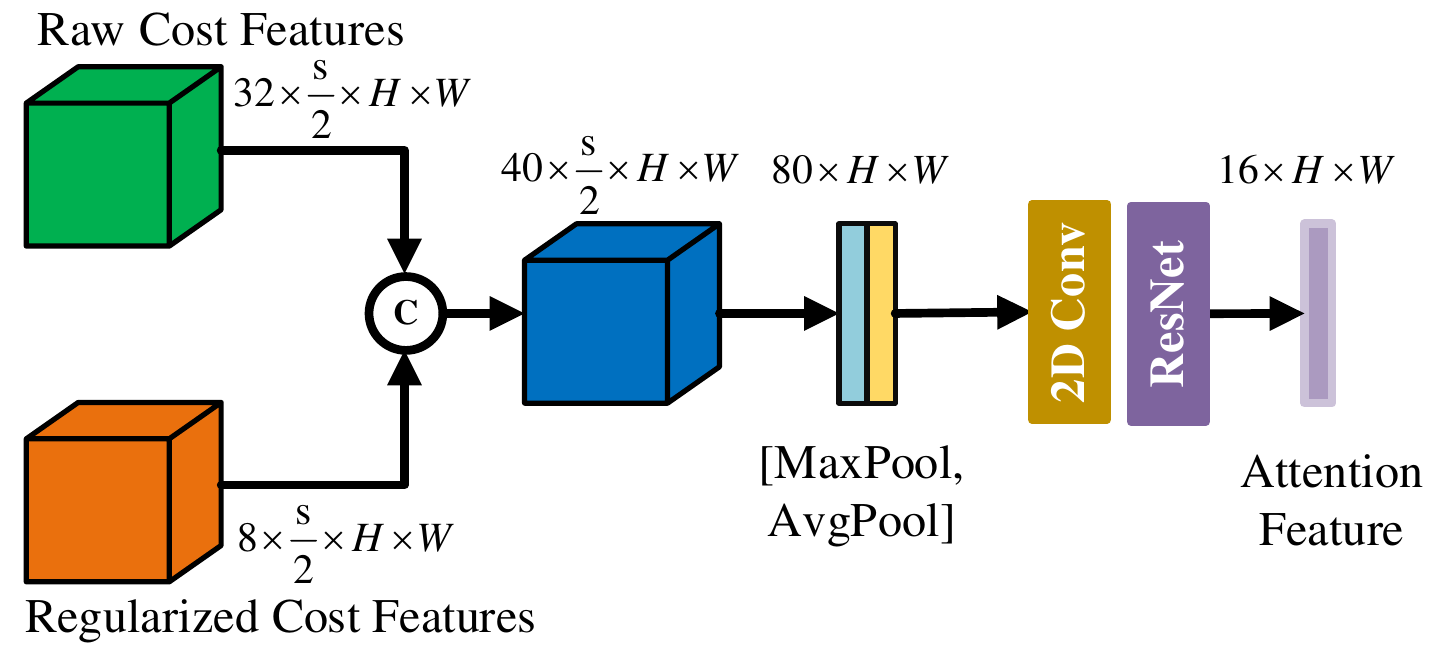}
	\caption{Structure of the depth attention module.}
	\label{fig:DA}
\end{figure}

The structure of the depth attention module is illustrated in Fig.~\ref{fig:DA}. 
For each block, we only use every other depth value in the block to sample raw cost features $\boldsymbol{F}_\text{raw}\in\mathbb{R}^{32\times \frac{s}{2}\times H \times W}$ and regularized cost features $\boldsymbol{F}_\text{reg}\in\mathbb{R}^{8\times \frac{s}{2}\times H \times W}$, where $H$ and $W$ denote the image height and width respectively. 
This enables the larger receptive fields in the depth direction.
We first concatenate these two kinds of cost features into complex cost features $\boldsymbol{F}_\text{comp}\in\mathbb{R}^{40\times \frac{s}{2}\times H \times W}$.
Such dense connection is good to explore their potential interactions and learn non-local information~\cite{Huang2017Densely}.
Then we use max-pooling and average-pooling along the depth dimension to aggregate discriminative depth information, generating two different depth context features, $\boldsymbol{F}_\text{max}$ and $\boldsymbol{F}_\text{avg}$.
As pointed out in \cite{Woo2018Cbam}, we think that max-pooling is good to gather distinctive clues while average-pooling is beneficial to learn the extent of depth space.
Using both can highlight the informative depth values as much as possible.
These two features are further concatenated and forwarded into one convolution layer and a residual block to produce attention features $\boldsymbol{F}_\text{att}\in\mathbb{R}^{16\times H \times W}$.

To further model the global context information in the depth direction, the attention features between blocks are interacted by another recurrent neural network.
Specifically, we reshape raw cost features $\boldsymbol{F}_\text{raw}\in\mathbb{R}^{32\times \frac{s}{2}\times H \times W}$ to $\mathbb{R}^{C \times H \times W}$, where $C=32\times \frac{s}{2}$. 
The input gate map ${\bf G}_i(t)$ and the forget gate map ${\bf G}_f(t)$ are modeled as
\begin{align}
\!\!\!\!\!{\bf G}_i(t)&=\sigmoid({\bf W}_{ia}\!\ast\![\boldsymbol{F}_\text{raw}(t),\boldsymbol{B}(t-1)]\!+{\bf b}_{ia}),\!\!\\
\!\!\!\!\!{\bf G}_f(t)&=\sigmoid({\bf W}_{fa}\!\ast\![\boldsymbol{F}_\text{raw}(t),\boldsymbol{B}(t-1)]\!+{\bf b}_{fa}),\!\!
\end{align}
where ${\bf W}$ is a transformation matrix and ${\bf b}$ is a bias term. 
Finally, we update the cell state map of the current block 
\begin{equation}
{\boldsymbol B}(t)={\bf G}_{i}(t)\odot\tanh(\boldsymbol{F}_\text{att}(t))+{\bf G}_{f}(t)\odot\boldsymbol{B}(t-1).
\end{equation}
By updating the cell states of different blocks in a gated recurrent manner, our method is able to capture global context information in the depth direction.
These cell states of blocks are in turn used to update the cell states of the LSTM cells.

\subsection{Loss Function}
%
Following previous practices~\cite{Yao2019RMVSNet,Yan2020Dense}, we cast the depth inference task as a multi-class classification problem.
We use the following cross-entropy loss to train our network:
\begin{equation}
{\cal{L}}=\sum_{\boldsymbol{p}\in\Phi}\sum_{d=0}^{D-1}-\boldsymbol{G}(d,\boldsymbol{p})\cdot\log(\boldsymbol{P}(d,\boldsymbol{p})),
\end{equation}
where $\Phi$ denotes the set of valid ground truth pixels, $\boldsymbol{G}(d,{\bf p})$ is the one-hot vector generated according to the ground truth depth map at pixel ${\bf p}$, and $\boldsymbol{P}(d,{\bf p})$ is the predicted depth probability at pixel ${\bf p}$.
Note that, we do not need to store the whole classification probability volume during testing.
Moreover, since the cell states of blocks are also sequentially updated, our designed depth attention module will not occupy too much memory. 

\subsection{Dynamic Depth Map Fusion}
%
After estimating depth maps for all input images, it is necessary to filter out wrong depth estimates for each depth map and then fuse them into a consistent point cloud representation.
There are two critical factors for filtering wrong depth predictions: depth probability and depth consistency.
The depth probability is usually measured by the corresponding probability of the selected depth, \emph{i.e.}, $\theta$.
The depth consistency is usually measured by the geometric constraint: reprojection errors and relative depth errors~\cite{Yao2018MVSNet}.
For a pixel ${\bf p}$ in the reference image $\boldsymbol{I}_0$, we denote its reprojection error w.r.t the source image $\boldsymbol{I}_i$ by $\psi_i$ and its relative depth error w.r.t. $\boldsymbol{I}_i$ by $\phi_i$.
Then, its consistency view set is defined as:
\begin{equation}
\mathcal{S}=\{\boldsymbol{I}_i|\psi_i<\epsilon,\phi_i<\eta\},
\end{equation}
where $\epsilon$ and $\eta$ are related thresholds.
If $|\mathcal{S}|>\mu$ and $\theta>\tau$, the depth estimate of ${\bf p}$ is reliable, where $\mu$ is the threshold of consistent view number and $\tau$ is the probability threshold.
Previous methods~\cite{Yao2018MVSNet,Yao2019RMVSNet,Gu2019Cas} usually set fixed threshold parameters to remove unreliable depth estimates.
However, these intuitively preset parameters make their fusion methods not robust for different scenes.
Thus, $\text{D}^2$HC-RMVSNet~\cite{Yan2020Dense} proposes a dynamic consistency checking algorithm to measure depth consistency.
Its dynamic consistency view set is defined as:
\begin{equation}
\mathcal{S}_d=\{\boldsymbol{I}_i|\psi_i<\epsilon(\mu),\phi_i<\eta(\mu),\epsilon(\mu)=\frac{\mu}{4},\eta(\mu)\!=\!\frac{\mu}{1300}\},
\end{equation}
The above definition makes the thresholds of geometric constraint be adaptively adjusted based on the consistent view number $\mu$. 
This means that the estimated depth value is accurate and reliable when satisfying strict depth consistency in a small number of views, or relaxed depth consistency in the majority of views.
However, this method still adopts a fixed depth probability threshold to filter depths with different depth consistency.
Due to the uncertainty of depth prediction networks~\cite{Kendall2017Uncertainties}, the depth probability will sometimes filter credible depth values that meet the strict depth consistency, making reconstructed point clouds less complete.
Thus, the depth probability threshold should also be adjusted according to the level of depth consistency. 
Based on these observations and our experiments, we define the dynamic probability threshold as:
\begin{equation}\label{Eq:dp}
\tau(\mu)=0.6\cdot \exp\left[{\frac{(\mu-10)}{8}}\right].
\end{equation}
By traversing different $\mu$ values, the estimated depth of ${\bf p}$ will be deemed accurate and reliable if $|\mathcal{S}_d|>\mu$ and $\theta>\tau(\mu)$.
That is, when the reprojection error and relative depth error are strict, the number of consistent view and depth probability threshold are relaxed to judge if the depth prediction is reliable.
Finally, the reliable depth estimates will be projected into 3D space to generate 3D point clouds.

\begin{table}[t]
	\centering
	\footnotesize
	\begin{tabular}{clccc}
		\toprule
		& {Method} & Acc.$\downarrow$ & Comp.$\downarrow$ & Overall$\downarrow$ \\
		\midrule
		\multirow{4}{0.04cm}{\rotatebox{90}{Geometric}} 
		& Furu & 0.613 & 0.941 & 0.777 \\
		& Tola & 0.342 & 1.190 & 0.766 \\
		& Gipuma & {\bf 0.283} & 0.873 & 0.578 \\
		& COLMAP & 0.400 & 0.664 & 0.532 \\
		\midrule
		\multirow{9}{0.04cm}{\rotatebox{90}{Learning-based}} & SurfaceNet & 0.450 & 1.040 & 0.745 \\
		& MVSNet & 0.396 & 0.527 & 0.462 \\
		& R-MVSNet & 0.383 & 0.452 & 0.417 \\ 
		& CasMVSNet & 0.325 & 0.385 & 0.355 \\
		& CVP-MVSNet & \underline{0.296} & 0.406 & \underline{0.351} \\
		& UCSNet & 0.338 & 0.349 & {\bf 0.344} \\
		& AttMVS & 0.391 & \underline{0.345} & 0.368 \\
		& $\text{D}^2$HC-RMVSNet & 0.395 & 0.378 & 0.386 \\
		& Ours & 0.370 & {\bf 0.332} & \underline{0.351} \\ 
		\bottomrule
	\end{tabular}
	\caption{Quantitative results on the DTU evaluation set using the distance metric [$mm$] (lower is better). Our method achieves the best completeness and the second best overall score. The best and second best results are in \textbf{bold} and \underline{underlined}, respectively.}
	\label{tab:dtuevaluation}
\end{table}  

\begin{figure}[t]
	\centering
	\includegraphics[width=\linewidth]{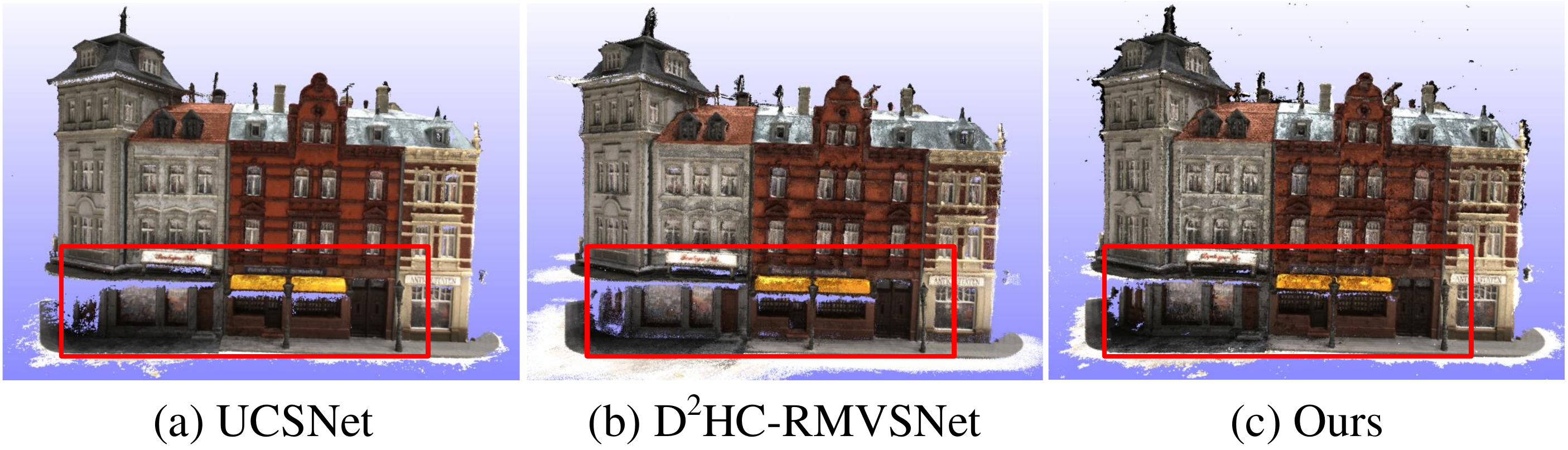}
	\caption{Qualitative results of reconstructed point clouds for the model scan 9 from the DTU evaluation set.}
	\label{fig:dtu}
\end{figure}

\section{Experiments}
%
We first describe datasets and implementation details. 
Then, we show benchmark results on the DTU dataset~\cite{Aanes2016Large} as well as the Tanks and Temples dataset~\cite{Knapitsch2017TTB}.
Finally, we conduct ablation studies to analyze our proposed core components.

\begin{table}[t]
	\centering
	\footnotesize
	\begin{tabular}{clccc}
		\toprule
		& Method & Acc.$[\%]\uparrow$ & Comp.$[\%]\uparrow$ & $F_1 [\%]\uparrow$ \\
		\midrule
		\multirow{3}{0.04cm}{\rotatebox{90}{Geo.}}  & COLMAP & 43.16 & 44.48 & 42.14 \\
		& ACMM & 49.19 & 70.85 & 57.27 \\
		& ACMP & 49.06 & \underline{73.58} & 58.41 \\
		\midrule
		\multirow{8}{0.04cm}{\rotatebox{90}{Learning-based}} & MVSNet & 40.23 & 49.70 & 43.48 \\
		& R-MVSNet & 43.74 & 57.90 & 48.40 \\
		& CVP-MVSNet & 51.41 & 60.19 & 54.03 \\
		& UCSNet & 46.66 & 70.34  & 54.83 \\
		& CasMVSNet & 53.71 & 63.88 & 56.84 \\
		& $\text{D}^2$HC-RMVSNet & 49.88 & \bf 74.08 & 59.20 \\
		& AttMVS & \bf 61.89 & 58.93 & \underline{60.05} \\
		& Ours & \underline{55.72} & 70.28 & \bf 60.49 \\
		\bottomrule
	\end{tabular} 
	\caption{Quantitative results on the Tanks and Temples Intermediate set. Our method achieves the best $F_1$-score. The best and second best results are in \textbf{bold} and \underline{underlined}.}
	\label{tab:t2evaluation}
\end{table}

\begin{figure}[t]
	\centering
	\includegraphics[width=0.99\linewidth]{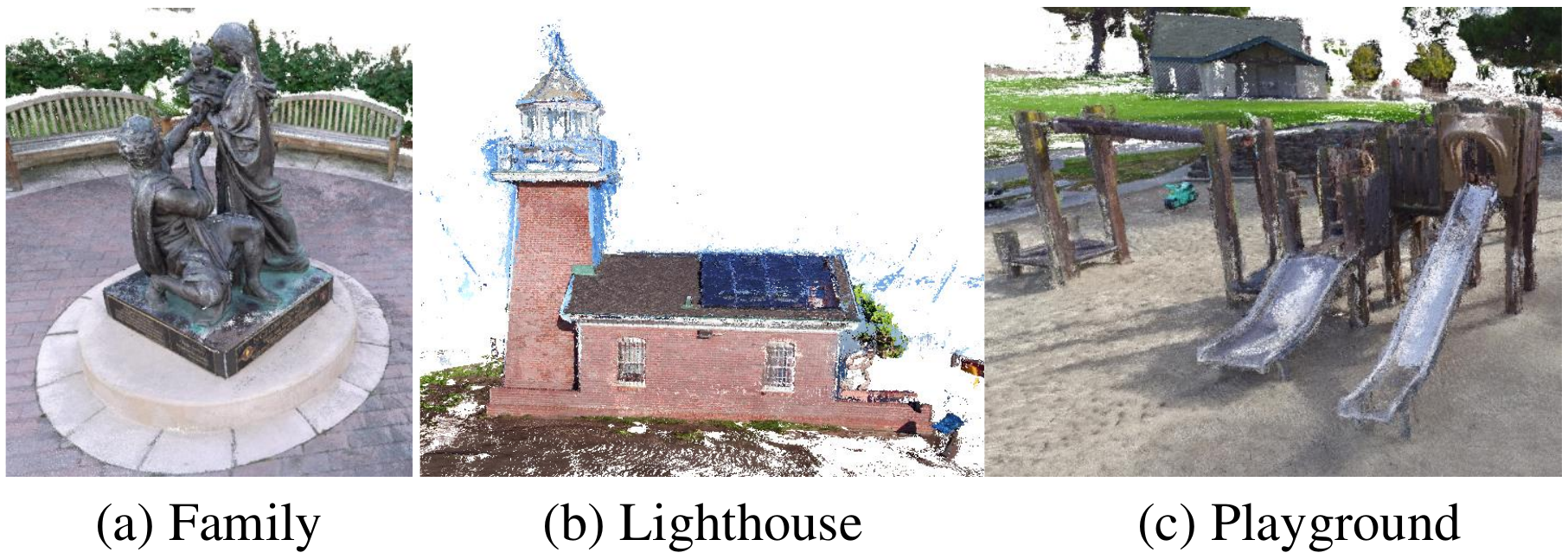}
	\caption{Qualitative results of reconstructed point clouds on Tanks and Temples dataset.}
	\label{fig:t2}
\end{figure}

\subsection{Datasets and Evaluation Metrics}
%
\boldparagraph{DTU dataset.} This dataset is captured from $49$ or $64$ views under a controlled indoor environment. 
$124$ scenes are included with $7$ different lighting conditions.
The ground truth camera parameters and 3D point clouds are both provided.
Following \cite{Ji2017Surfacenet,Yao2018MVSNet}, we divide this dataset into training, validation and evaluation sets.

\boldparagraph{Tanks and Temples dataset.} This dataset contains both indoor and outdoor scenes, which are captured in the realistic environment.
Moreover, unlike the object-centric scenes in DTU, these scenes are larger and more complex.
This dataset is divided into Intermediate set and Advanced set. 
The latter one is more challenging due to scale, textureless regions and complicated visibility. 

\boldparagraph{Evaluation metrics.} To evaluate the performance of our method on different datasets, distance metric and percentage metric are used for the DTU and Tanks and Temples dataset, respectively.
For the distance metric, the overall score defined as the mean of accuracy and completeness measures the overall performance of reconstructed point clouds.
For the percentage metric, the $F_1$ score defined as the harmonic mean of accuracy and completeness is adopted.   

\subsection{Implementation Details}\label{Sec:Id}
%
\boldparagraph{Training.} We train our network with ground truth depth maps on the DTU training set.
The ground truth depth maps are generated through Poisson Surface Reconstruction~\cite{Kazhdan2013SPS}.
Due to the limitation of GPU memory, the input images are resized to $H\times W=128\times 160$.
The input image number is $N=7$.
The depth values are sampled from $425mm$ to $905mm$ with $D=192$ in an inverse depth manner.
The block size $s$ is set to $8$.
Our network is implemented using Pytorch~\cite{NIPS2019PyTorch} and trained with Adam~\cite{Kingma2015Adam} optimizer end-to-end for $10$ epochs on two NVIDIA RTX 2080Ti GPU cards.
The initial learning rate is set to $0.001$ and the batch size is set to $2$.
Note that, to prevent estimated depth maps from being biased on the recurrent regularization order, we randomly use forward or backward pass to train samples.

\boldparagraph{Evaluation.} During the evaluation, we only use the forward pass to infer depth maps.
We sample $D=512$ depth planes in an inverse depth manner~\cite{Yao2019RMVSNet,Xu2020Learning} and set the input image number $N$ as 7.
For the DTU dataset, the input image resolution is set to $H\times W=600\times 800$. 
For the Tanks and Temples dataset, we use the camera parameters provided by R-MVSNet~\cite{Yao2019RMVSNet} to evaluate our method with the image resolution $H\times W=544\times 960$.
Following previous practices~\cite{Yao2019RMVSNet,Yan2020Dense}, we first use the maximum classification probability as the depth probability of estimated depths.
Then, we apply the designed dynamic depth map fusion method to generate the final point clouds.

\subsection{Benchmarking Results}
%
In this section, we directly use our model trained on the DTU training set without any fine-tuning for benchmarking evaluations.
We compare our method with other state-of-the-art MVS methods, including geometric and learning-based methods. 
The geometric methods include Furu~\cite{Furukawa2010Accurate}, Tola~\cite{Tola2012Efficient}, Gipuma~\cite{Galliani2015Massively}, COLMAP~\cite{Schonberger2016Pixelwise}, ACMM~\cite{Xu2019Multi} and ACMP~\cite{Xu2020Planar}. For the learning-based methods, SurfaceNet~\cite{Ji2017Surfacenet}, MVSNet~\cite{Yao2018MVSNet}, R-MVSNet~\cite{Yao2019RMVSNet}, CasMVSNet~\cite{Gu2019Cas}, CVP-MVSNet~\cite{Yang2020CVPMVSNet}, UCSNet~\cite{Cheng2020UCSNet}, AttMVS~\cite{Luo2020Attention} and $\text{D}^2$HC-RMVSNet~\cite{Yan2020Dense} are compared.

\boldparagraph{Results on DTU.} We evaluate our method on the DTU evaluation set. 
The comparison results are shown in Table~\ref{tab:dtuevaluation}.
Our method achieves the best completeness and competitive overall performance among compared methods. 
In particular, our method outperforms the previous recurrent regularization methods, including R-MVSNet and $\text{D}^2$HC-RMVSNet.
Moreover, our method is very competitive with the 3D filtering methods, \emph{e.g.}, CVP-MVSNet and UCSNet. 
The qualitative comparisons in Fig.~\ref{fig:dtu} show that our reconstructed point clouds are more complete than UCSNet and $\text{D}^2$HC-RMVSNet. 

\boldparagraph{Results on Tanks and Temples.} We evaluate the generalization ability of NR2-Net on the Tanks and Temples dataset. 
Table~\ref{tab:t2evaluation} summarizes the evaluation results.
As can be seen, our method yields the best mean $F_1$ score, $60.49\%$, over all published methods, demonstrating that our method generalizes better than other methods.
In addition, our method achieves much better performance than other MVS methods based on 3D cost volume filtering on this dataset.
We believe this is because the scenes in Tanks and Temples are larger than the scenes in DTU, while the 3D cost volume filtering methods can only sample limited depth planes due to the GPU memory limitation.
In contrast, our method can sample sufficient depth values and better considers the long-range dependencies in the depth direction, making it more feasible in practice. 
Fig.~\ref{fig:t2} shows the qualitative results of our reconstructed point clouds.


\begin{table}[t]
	\centering
	\footnotesize
	\begin{tabular}{l|cc|ccc}
		\toprule
		Model & NLR & DF  & Acc.$\downarrow$ &  Comp.$\downarrow$ & Overall$\downarrow$ \\
		\midrule
		Baseline & & & 0.350 & 0.473 & 0.411 \\
		Model-A & & $\checkmark$ & 0.378 & 0.405 & 0.391 \\
		Model-B & $\checkmark$ & & \bf 0.344 & 0.364 & 0.354 \\
		Full & $\checkmark$ & $\checkmark$ & 0.370 & \bf 0.332 & \bf 0.351 \\ 
		\bottomrule
	\end{tabular}
	\caption{Ablation study of different components in our NR2-Net. NLR and DF mean non-local recurrent regularization and dynamic depth map fusion respectively.}
	\label{tab:ablation}
\end{table}

\begin{figure}[t]
	\centering
	\includegraphics[width=\columnwidth]{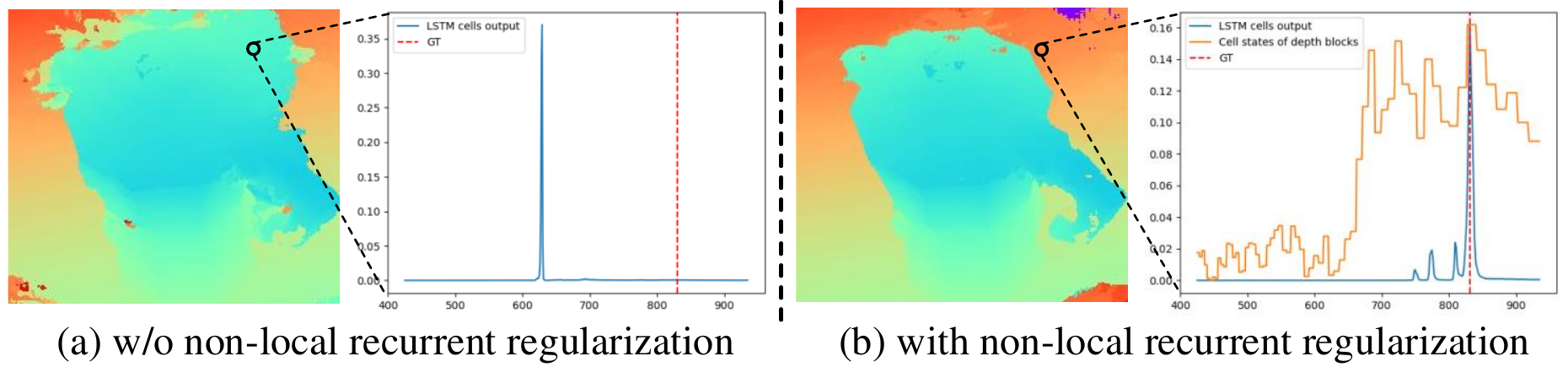}
	\caption{Visual comparison of predicted probability distribution at one pixel between Baseline (a) and Full model (b). The cell states of depth blocks indicate the circle region belongs to the background and constrain the probability distribution of cost map regularization to produce true estimates.}
	\label{fig:vis}
\end{figure}

\begin{table}[t]
	\centering
	\footnotesize
	\begin{tabular}{lcccc}
		\toprule
		Method & Input Size & Output Size & Mem. & Time \\
		\midrule
		MVSNet & 800$\times$576 & 200$\times$144 & 8.9 & 0.78 \\
		R-MVSNet & 800$\times$576 & 200$\times$144 & 1.2 & 0.89 \\
		D$^2$HC-RMVSNet & 200$\times$144 & 200$\times$144 & 0.9 & 2.67  \\
		Ours & 200$\times$144 & 200$\times$144 & 1.0 & 1.90 \\
		\midrule
		CVP-MVSNet & 800$\times$576 & 800$\times$576 & 2.2 & 0.49 \\
		D$^2$HC-RMVSNet & 800$\times$576 & 800$\times$576 & 2.4 & 20.94 \\
		Ours & 800$\times$576 & 800$\times$576 & 3.4 & 16.65 \\
		\bottomrule
	\end{tabular}
	\caption{Evaluation of GPU memory [GB] and runtime [s] for different methods.}
	\label{tab:TM}
\end{table}

\subsection{Ablation Study}
%
\boldparagraph{Effect of different components.} In this section, we conduct an ablation study on the DTU evaluation set to verify the effectiveness of non-local recurrent regularization and dynamic depth map fusion module in the proposed NR2-Net.
The baseline model is created by removing the non-local recurrent regularization module and replacing the dynamic depth map fusion with dynamic consistency checking, which means that Eq.~\eqref{Eq:dp} becomes $\tau(\mu)=0.35$ as \cite{Yan2020Dense} suggested.
The baseline model is also trained with the procedure as described in the previous section. 
Based on the baseline model, we add the non-local recurrent regularization module and use the dynamic depth map fusion step by step to construct model-A, Model-B and Full model. The comparison results are listed in Table~\ref{tab:ablation}.
This clearly demonstrates the effectiveness of our proposed two core modules in NR2-Net. In particular, by comparing Baseline and Model-B, Model-A and Full model, we observe that our proposed non-local regularization improves accuracy and greatly boosts completeness, leading to significant overall performance improvement.
By comparing Baseline and Model-A, Model-B and Full model, we see that our proposed dynamic fusion can produce more complete point clouds, resulting in better overall performance.
This also demonstrates the generalization of our proposed fusion for different depth inference networks.

To further demonstrate the effect of the proposed non-local recurrent regularization, we try to visualize the cell states of different depth blocks.
In fact, it is hard to give comprehensive visualization about these cell states. Instead, we show the probability distribution of these cell states to see whether they indicate the global scene context.
To this end, we average the features of $\boldsymbol{B}(t)$ along the channel dimension and apply the softmax operation along the depth dimension for all depth blocks to obtain the probability distribution.
Then, this probability distribution is aligned with the probability distribution of all depth planes by copying the value of each block to its covering $s$ depth planes.
The visualization results are shown in Fig.~\ref{fig:vis}.
The circle region in Fig.~\ref{fig:vis} is challenging as it is close to the central object in the spatial domain but in fact it belongs to the background areas.
Without non-local recurrent regularization, it is easy for the baseline model to propagate the depth information of the central object to the circle region, resulting in incorrect estimation for this region (cf. Fig.~\ref{fig:vis}(a)).
With our proposed non-local recurrent regularization, the cell states of depth blocks indicate the circle region belongs to the background areas and constrain the regularization of LSTM cells (cf. Fig.~\ref{fig:vis}(b)).
This demonstrates to some extent that the cell states of depth blocks capture the global scene context to help cost map regularization.  

\boldparagraph{Runtime and GPU memory.} We evaluate the runtime and GPU memory usage of our method to generate each depth map on the DTU evaluation set. 
Table~\ref{tab:TM} shows the comparison results with other methods. 
For pair comparison with MVSNet and R-MVSNet, we run different methods with the same depth sample number $D=256$.
We see that 3D filtering methods take up a lot of memory and cannot tackle high-res input.
Recurrent regularization methods greatly reduce memory consumption at the cost of runtime and can tackle high-res input.
Although cascade 3D filtering methods such as CVP-MVSNet employ the coarse-to-fine strategy to fit in high-res input, they can only sample limited depth planes in each stage.
This prevents them from perceiving full-space global context information in the depth direction, hindering their performance in practical large-scale scenes, \emph{e.g.}, Tanks and Temples dataset.

\section{Conclusion}
%
We presented a novel non-local recurrent regularization network for multi-view stereo. 
A novel depth attention module is proposed to model non-local interactions within depth blocks. 
These non-local interactions are updated in a gated recurrent way to model global scene context. 
In this way, long-range dependencies along the depth direction are captured and in turn utilized to help regularize cost maps.
In addition, we design a dynamic depth map fusion strategy to adaptively take both the depth probability and depth consistency into account. 
This further improves the robustness of the point cloud reconstruction. 
Ablation studies demonstrate the effectiveness of our proposed core components. 
Experimental results show that our method achieves competitive results on the indoor DTU dataset and exhibits excellent performance on the large-scale Tanks and Temples dataset. 

\bibliography{refs}

\end{document}